\documentclass[letterpaper, 10 pt, conference]{ieeeconf}  % Comment this line out if you need a4paper

\usepackage{cite}
\usepackage{amsmath,amssymb,amsfonts}
\usepackage{graphicx} 
\usepackage{xcolor}
\usepackage{subcaption}
\PassOptionsToPackage{hyphens}{url}
\usepackage{comment}

\makeatletter
\let\NAT@parse\undefined
\makeatother
\usepackage{hyperref}

\usepackage{caption}
\usepackage{stfloats}
\usepackage{booktabs}

\usepackage{tikz}
\usetikzlibrary{shapes.geometric, arrows, positioning}

\usepackage{siunitx}

\definecolor{darkgreen}{RGB}{0,164,0}

\tikzstyle{decision} = [diamond, draw, fill=blue!20, 
text width=4.5em, text badly centered, inner sep=0pt, font=\small]
\tikzstyle{process} = [rectangle, draw, fill=orange!30, 
text width=5em, text centered, rounded corners, minimum height=4em, font=\small]
\tikzstyle{line} = [draw, -latex']

\title{\LARGE \bf
Real-Time Bounded Catenary Solver for UAV Tether Modeling
}

\author{Max Beffert and Andreas Zell\\
Cognitive Systems Group, University of Tübingen, Germany\\
\texttt{max.beffert@uni-tuebingen.de}
}

\newcommand{\refappendix}[1]{%
	\hyperref[#1]{Appendix~\ref*{#1}}%
}

\begin{document}

\maketitle
\thispagestyle{empty}
\pagestyle{empty}

%\AddToHookNext{shipout/background}{\copyrightnotice}

%%%%%%%%%%%%%%%%%%%%%%%%%%%%%%%%%%%%%%%%%%%%%%%%%%%%%%%%%%%%%%%%%%%%%%%%%%%%%%%%

\begin{abstract}
	For non-stationary tethered multirotor UAVs in real-world conditions, simulating the forces imposed on the drone by the aerodynamic drag of the tether becomes crucial, with online use cases placing a hard bound on the maximum solve time. 
	In previous work, a quasi-analytical catenary tether model reached a mean solve time of \qty{0.51}{\milli\second} using a general-purpose root finder, but without any worst-case guarantees or proven convergence.
	In this work, we reformulate the inner solver by reducing the catenary boundary-value problem to a single transcendental equation in one well-conditioned unknown. We derive a closed-form bracket and prove monotonicity and convexity as well as existence and uniqueness of the root, which together guarantee convergence of the solver. We further propose a two-regime initial guess which approximates the true root within \qty{3.4}{\percent} and reduces the mean iteration count by \qty{69.2}{\percent} to \num{2.24} compared to the textbook initialization.
	Building on the hybrid root-finding method \emph{rtsafe} (Newton-Raphson with bisection fallback giving bounded iteration counts), we implement a specialized variant that exploits the problem structure to omit unnecessary checks while retaining correctness, which gives up to \num{1.37} times speedup.
	With the proposed solver the full tether model achieves a nearly constant solve time of \qty{4.0}{\micro\second} on average and \qty{4.6}{\micro\second} at worst, a \num{68} times speedup over an optimized re-implementation of the previous method, while agreeing with it to a relative deviation of \num{8.7e-9}. Because the reformulation leaves the underlying physical model untouched, the experimental validation of the previous work carries over unchanged. We further demonstrate its suitability for embedded, resource-constrained platforms with a Lua implementation running directly in ArduPilot on a drone's flight controller, where it stays well inside the scheduling budget with a mean solve time of \qty{0.74}{\milli\second}.
\end{abstract}

%%%%%%%%%%%%%%%%%%%%%%%%%%%%%%%%%%%%%%%%%%%%%%%%%%%%%%%%%%%%%%%%%%%%%%%%%%%%%%%%
\section{Introduction}

A fundamental limitation of multirotor UAVs is their short flight time compared to helicopters and fixed-wing aircraft due to their lower efficiency. One promising approach to achieve continuous operation is to power the drone from the ground through a tether. This is already implemented in some practical applications but mainly for static operation in low wind scenarios. The reason for this limited scope is that the cable aerodynamics become relevant in cases with higher airspeed due to motion or wind. For precise system design, control, trajectory planning, and simulation, it is therefore crucial to estimate or measure the forces the tether imposes on the drone due to its weight and aerodynamic drag. In previous work \cite{beffert_low-latency_2026}, two complementary quasi-static tether models were proposed for this purpose, an analytical one based on catenary theory and a numerical one that discretizes the tether into segments and lumped masses. Both were validated against tension measurements on a real-world flight and shown to have close agreement with each other and the measurements.

The analytical model is the fast path of that framework and the one intended for use in real-time scenarios, for example inside a control or estimation loop. However, the solver at its core had no convergence guarantees or upper bounds on runtime. It achieved an acceptable mean solve time of \qty{0.51}{\milli\second}, but with a high variance, making it unsuitable for real-time use. Furthermore, it is questionable whether the solve speed and implementation would transfer well to the low computational budget of the drone's flight controller.

The contributions of this work are:
\begin{itemize}
	\item reduction of the catenary problem to a single transcendental equation in one well-conditioned unknown, with closed-form recovery of the remaining parameters,
	\item proof of guaranteed convergence and bounded iteration count,
	\item an initial guess that reduces the mean iteration count by \qty{69.2}{\percent} compared to the textbook initialization,
	\item a specialized solver achieving up to \num{1.37} times speedup over textbook \emph{rtsafe},
	\item an evaluation on real flight data, with a nearly constant runtime showing a \num{68} times speedup over an optimized re-implementation of the previous method,
	\item a Lua implementation running on the drone's flight controller with limited resources.
\end{itemize}

\section{Related Work}
For a broader overview of tether modeling approaches refer to \cite{beffert_low-latency_2026}. Here we recall only what is needed to place the present contribution, and refer to Table~\ref{ComparisonTable} for the overall positioning compared to other analytical and quasi-analytical methods. Purely analytical catenary models \cite{borgese_tether-based_2022, jain_tethered_2022} are fast but neglect aerodynamic forces entirely. Borgese et al. \cite{borgese_tether-based_2022} avoid an implicit solve altogether by measuring the tether angle directly at both ends and recovering the shape parameter in closed form; we therefore omit it from Table~\ref{ComparisonTable}, since no solve is involved. Jain et al. \cite{jain_tethered_2022} compute the tether shape between a series of drones. They state that the catenary parameters are solved numerically without specifying the method or reporting solve time or convergence behavior.

Quasi-analytical models add drag under a uniformity assumption \cite{schmehl_analytical_2018,beffert_low-latency_2026}. Bigi et al. \cite{schmehl_analytical_2018} solve the inner catenary shape with a fixed-point iteration and state that it converges for all possible values. They mention a runtime of below \qty{1}{\second} for the whole approach, but how much of that is taken up by the catenary solve is not stated, and no iteration bounds are given.

Where the catenary boundary-value problem is judged too costly to solve online, some works avoid it differently: Talke et al. \cite{talke_catenary_2018} note that the transcendental catenary equation admits no purely analytic solution and instead perform an offline brute-force sweep over a discretized grid of relative positions and tether lengths, fitting the resulting tension, length, and departure angle to low-order polynomials.

Numerical quasi-static formulations \cite{koenemann_modeling_2017, zanon_airborne_2014, beffert_low-latency_2026} and fully dynamic ones \cite{dicembrini_modelling_2020, muttin_umbilical_2011} resolve per-segment drag and, in the dynamic case, transient behavior, at a computational cost that grows with the discretization. Beffert et al. \cite{beffert_low-latency_2026} showed via real-world validation that quasi-analytical methods have close agreement with numerical quasi-static methods for common cases of tethered drones at reduced computational cost. 

Most of the literature does not target fast runtime, and solve time figures are therefore often not provided. Furthermore, we are not aware of any previous work that gives real-time guarantees such as a proven convergence and a bound on iteration count.

\subsection{Previous Method}
This work improves the analytical model proposed in \cite{beffert_low-latency_2026}, which rests on the observation that a quasi-static tether follows the catenary curve in a uniform potential field. By assuming that the aerodynamic load acts equally along the whole tether, the drag force and tether weight can be combined into a single unified potential field. Therefore, the shape follows the catenary curve, but in a frame rotated such that the potential field points downward. To obtain the tether shape, the endpoints are rotated into that frame, solved there, and the sampled shape and tensions are rotated back. 

The total drag depends on the exposed area of the cable, which in turn depends on the vertical length of the tether, which is itself an output of the solve. Therefore, the algorithm starts from the vertical distance $|\Delta y|$ (the case without any sag) and refines it iteratively by solving the shape and recomputing the vertical length until it converges. For scenarios typically observed with tethered drones, the first estimate is already close and the loop converges quickly.

At the core of the method sits a numerical solver that finds the catenary parameters that describe the cable shape for a given configuration of endpoint coordinates and tether length. The constraint equations ensure that the endpoints lie on the catenary and that the arc length matches the tether length. In \cite{beffert_low-latency_2026} this system of three coupled equations is handed to \emph{scipy.fsolve}, a general-purpose Newton-type root finder. To achieve successful solves over a wide range of conditions, multiple sets of initial guesses based on parabolic approximation and heuristics have to be tried until one succeeds. The cost of these guesses was somewhat counteracted by statistical analysis trying ones first that are more likely to succeed, but it remained a source of solve time variance. 

The approach was demonstrated to work in practice, but for the solver used no convergence guarantees or runtime bounds could be given. The present work replaces only the inner catenary solver with one where these guarantees can be given and proven but leaves the surrounding formulation, including the rotated-frame and drag-estimation outer loop, unchanged.

\begin{table*}[t]
	\centering
	\small
	\caption{Comparison of analytical tether modeling approaches. For real-time use, guaranteed convergence, bounded runtime, and low latency are desired.}
	\begin{tabular}{l l c c c c}
		\toprule
		Work & Method & Drag & Guaranteed Convergence & Bounded Runtime & Low-Latency \\
		\midrule
		
		Jain et al. \cite{jain_tethered_2022}
		& analytical 
		& \textcolor{red}{no} 
		& \textcolor{red}{not stated} 
		& \textcolor{red}{not stated} 
		& \textcolor{red}{not stated} \\
		
		Bigi et al. \cite{schmehl_analytical_2018}
		& quasi-analytical 
		& \textcolor{darkgreen}{yes} 
		& \textcolor{orange}{yes (no proof)} 
		& \textcolor{red}{no} 
		& \textcolor{orange}{not stated (\qty{1}{\second} full method)} \\
		
		Beffert et al. \cite{beffert_low-latency_2026}
		& quasi-analytical 
		& \textcolor{darkgreen}{yes} 
		& \textcolor{red}{no} 
		& \textcolor{red}{no} 
		& \textcolor{darkgreen}{yes (\qty{0.51}{\milli\second})} \\
		
		\textbf{Ours} 
		& \textbf{quasi-analytical} 
		& \textbf{\textcolor{darkgreen}{yes}} 
		& \textbf{\textcolor{darkgreen}{yes (proven)}} 
		& \textbf{\textcolor{darkgreen}{yes}} 
		& \textbf{\textcolor{darkgreen}{yes (\qty{4.0}{\micro\second})}} \\
		
		\bottomrule
	\end{tabular}
	\label{ComparisonTable}
\end{table*}

\begin{comment}
\begin{figure}[htbp]
	\centering
	\includegraphics[width=0.3\textwidth]{figures/Tether.jpg}
	\caption{A photo of the tethered drone showing the cable curvature.}
	\label{tether}
\end{figure}
\end{comment}

\section{Methods}

\subsection{Analytical Problem Formulation}

The cable follows a catenary curve given by
\begin{equation}
	y(x) = a \cosh\big((x - x_0)/a\big) + y_0.
\end{equation}
Therefore, the tether shape is fully determined by the scale parameter $a>0$ and the vertex coordinates $(x_0, y_0)$. Determining these three unknowns requires three constraints: both endpoints lie on the curve and the arc length between them is equal to the cable length $L$. As shown in \refappendix{analytical-appendix}, combining these equations and simplifying reduces the system to
\begin{equation}
	L^2 - \Delta y^2 = 4a^2 \sinh^2 b, \qquad b := \frac{\Delta x}{2a}.
	\label{eq:intermediate}
\end{equation}
Substituting the definition of $b$, taking the positive square root, and considering $a>0$ gives
\begin{equation}
	g(a) = 2a \, \sinh\left(\frac{|\Delta x|}{2a}\right) - \sqrt{L^2 - \Delta y^2} = 0.
	\label{eq:g_of_a}
\end{equation}
While \eqref{eq:g_of_a} is a standard reduction of the catenary problem \cite{beitelschmidt_catenary_2025}, it is numerically unfavorable for root-finding:
\begin{itemize}
	\item $a$ ranges over the unbounded interval $(0, \infty)$, which is unsuitable for bracketing methods
	\item the argument $|\Delta x|/2a$ diverges as $a\,\mathord{\to}\,0$, causing $\sinh(\cdot)$ to diverge
	\item as $a\,\mathord{\to}\,\infty$ the derivative of $g$ vanishes, stalling Newton-type solvers
\end{itemize}
Instead, we solve for the unsigned quantity $\hat b := |\Delta x| / 2a$, by inverting the substitution and eliminating $a$ instead of $b$:
\begin{equation}
	g(\hat b) = |\Delta x|\,\frac{\sinh \hat b}{\hat b} - \sqrt{L^2 - \Delta y^2} = 0,
	\label{eq:g_of_b}
\end{equation}
which matches \eqref{eq:g_of_b_appendix} in \refappendix{reduction-analytical-appendix}. Unlike $a$, $\hat b$ is well suited for root finding:
\begin{itemize}
	\item $\hat b$ is confined to a finite interval with analytical bounds
	\item no singularity: $\sinh(\hat b)/\hat b \to 1$ as $\hat b \to 0$
	\item the derivative of $g$ remains well-behaved over the whole domain
\end{itemize}
We therefore solve \eqref{eq:g_of_b} for $\hat b$ and recover $a$, $x_0$, and $y_0$ in closed form as detailed in \refappendix{recovery-analytical-appendix}.
For almost taut tethers, \eqref{eq:g_of_b} subtracts nearly equal numbers, which loses floating-point precision (catastrophic cancellation). We therefore solve an equivalent, cancellation-free form of $g(\hat b)$ (\refappendix{reduction-analytical-appendix}) and, for small $\hat b$, use the Taylor series of $\sinh(\hat b)/\hat b$ and its derivative, truncated after $\hat b^8$ and $\hat b^7$. The switching point lies where truncation and rounding errors balance: \num{0.15} in float64 and \num{1} in float32. Above the switching point, $g$ and $g'$ are evaluated together from a single exponential, so each evaluation needs only one function call.

\subsection{Well-posedness Guards}\label{sec:guards}
Two conditions must hold for \eqref{eq:g_of_b} to admit a root, and both are checked before the solver is entered. First, the cable must be strictly longer than the straight-line distance between the endpoints, $L > \sqrt{\Delta x^2 + \Delta y^2}\,(1+\varepsilon)$ with $\varepsilon = \num{e-6}$. A taut or over-stretched cable is outside the scope of the inextensible catenary model and is rejected rather than approximated.

Dividing \eqref{eq:g_of_b} by $|\Delta x|$ leaves $\sinh(\hat b)/\hat b = r$ with $r := \sqrt{L^2-\Delta y^2}/|\Delta x|$, so the geometry enters only through $r$. It measures excess length relative to horizontal span: $r \to 1$ is a taut cable, while large $r$ arises from slack, a small horizontal span, or both. The first condition keeps $r$ strictly above $1$. Second, the horizontal span must not be degenerate, since $\Delta x = 0$ gives $r \to \infty$. We therefore require $r \leq r_{\mathrm{max}} = \num{e6}$, which also bounds the bracket. The length condition is invariant under the frame rotation for drag and only needs to be tested once, but $r$ depends on $\Delta x$ in the rotated frame and is re-tested at every iteration of the outer loop.

\subsection{Solver}\label{sec:solver}

Equation~\eqref{eq:g_of_b} is solved using \emph{rtsafe}, a hybrid root-finding method introduced in Numerical Recipes~\cite{press_numerical_2007} that combines the guaranteed convergence of bisection with the fast local convergence of Newton's method. The algorithm maintains a bracket $[\hat b_{\mathrm{lo}}, \hat b_{\mathrm{hi}}]$ known to contain the root, and refines it iteratively. At each iteration, a candidate step is proposed using the Newton update 
\begin{equation}
	\hat b_{n+1} = \hat b_n - g(\hat b_n)/g'(\hat b_n). 
\end{equation}
This step is accepted only if it remains within the current bracket and shrinks it at a sufficient rate. Otherwise, the algorithm falls back to a bisection step, $\hat b_{n+1} = (\hat b_{\mathrm{lo}}+\hat b_{\mathrm{hi}})/2$. At the end of every iteration the function is evaluated at the new point to determine whether $\hat b_{n+1}$ becomes the new upper or lower bracket bound.

Because the bracket always contains the root (\refappendix{existence-appendix}) and its width strictly decreases, \emph{rtsafe} cannot diverge or oscillate the way standard Newton can when the starting point is poor. Convergence is therefore guaranteed.

For real-time use, it is also important to have an upper bound on how many iterations are necessary to converge. Let $\Delta_n := |\hat b_{n+1} - \hat b_n|$ denote the length of the step taken at iteration $n$, whether it was a Newton or a bisection step. A proposed Newton step is then accepted only if
\begin{equation}
	\left|\frac{g_n}{g'_n}\right| \leq \frac{\Delta_{n-2}}{2},
	\label{eq:stall_check_main}
\end{equation}
i.e., if it is at most half as long as the step taken two iterations earlier. The step length is therefore halved at least every two iterations. We terminate once the step falls below a tolerance relative to the root, so it can be chosen for the resolution of the floating-point format (Sec.~\ref{sec:tolerance}). As the root is unknown, we scale by the initial guess \eqref{eq:initial-guess} instead, which is a close approximation, and stop once $\Delta_n < \varepsilon_b\,\hat b_0$. This bounds the iteration count by $N_{\mathrm{max}} = 2\lceil \log_2\!\big(\hat b_{\mathrm{hi}}/(\varepsilon_b\,\hat b_0)\big) \rceil$.

Newton-Raphson by itself is already guaranteed to converge for $g(\hat b)$ since it is monotonic and convex (\refappendix{monotonicity-appendix}) \cite{thorlund-petersen_global_2004}, but without any guarantees of how many iterations are necessary. What \emph{rtsafe} adds is an upper bound for the iterations needed to converge. This makes it suitable for real-time applications, where bounded worst-case runtime is crucial, not just the average solve time.

Solving the catenary with \emph{rtsafe} requires providing valid bounds for $\hat b$. As previously discussed, $\hat b$ is positive, so $0$ is used as the lower bound. The upper bound is determined analytically (\refappendix{bounds-appendix}) in the cancellation-free form:
\begin{equation}\label{eq:upper-bound}
	\hat b_{\mathrm{hi}}=\sqrt{120(r-1)\big/\big(10 + \sqrt{100 + 120(r-1)}\big)}.
\end{equation}
This bound stays close to the true root, so the number of bisection steps compared to a perfect bracket is below half a step for $r \leq \num{100}$ and below three even at $r_{\mathrm{max}}$ (\refappendix{bounds-tightness-appendix}).

In standard \emph{rtsafe}, the middle of the bracket is used as the initial guess for the first Newton step. This works, but is not optimal, since Newton-Raphson only converges quadratically close to the root, so the quality of the guess strongly impacts solve time. Instead, an analytical guess is used to initialize the first Newton step. For small $\hat b$, the upper bound \eqref{eq:upper-bound} serves as an accurate initial guess; for large $\hat b$, we derive a Lambert $W$ approximation in \refappendix{initial-guess-appendix}, with $\Lambda := \ln(2r)$. The optimal regime boundary at \num{3.77} was determined numerically (Sec.~\ref{sec:guess-exp}).
\begin{equation}
		\hat b_0 =
	\begin{cases}
				\hat b_{\mathrm{hi}}, & r < 3.77,\\[2pt]
		\Lambda + \ln \Lambda + \ln \Lambda/\Lambda, & r \geq 3.77.
	\end{cases}
	\label{eq:initial-guess}
\end{equation}
Since the Lambert $W$ approximation is not guaranteed to underestimate the root, we clamp the guess to $[0, \hat b_{\mathrm{hi}}]$.

While the problem can be solved with textbook \emph{rtsafe}, it is possible to remove some unnecessary checks by using specific function knowledge. The main optimization is removing the bracket validation, which confirms that a sign change occurs within the bracket in standard \emph{rtsafe}. In this case the bracket is always valid because it is calculated analytically (\refappendix{bounds-appendix}). This is significant since the bracket validation requires two costly evaluations of $g(\hat b)$.

Furthermore, when deciding whether a Newton step can be accepted, checking whether it falls below the lower bracket is not necessary. This follows from the monotonic convex nature of the function and is shown in detail in \refappendix{low-bracket-appendix}. This is not an expensive check, but it happens every iteration, so the savings add up with higher iteration counts.

\section{Experiments and Results}

All desktop measurements were taken on an Intel Core i7-9700 CPU. The embedded results come from a Pixhawk~6C flight controller running ArduPilot V4.6.3. We use two real flights, \emph{Flight~1} from \cite{beffert_low-latency_2026}, which comprises \num{170} tether configurations (\num{266} inner-loop solves) with $r$ ranging from \num{1.050} to \num{2724} (median \num{2.0}), and \emph{Flight~2}, which comprises \num{4875} configurations.

\subsection{Initial Guess}
\label{sec:guess-exp}

\begin{figure}[tb]
	\centering
	\includegraphics[width=\columnwidth]{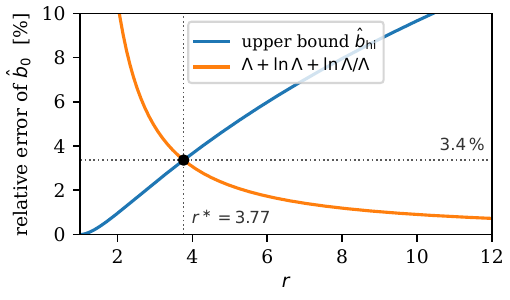}
	\caption{Relative error of the two initial guesses \eqref{eq:initial-guess} compared to the actual root. The curves cross at $r^\ast = \num{3.77}$ at a relative error of \qty{3.4}{\percent}, which is where we place the regime boundary.}
	\label{fig:initial-guess-error}
\end{figure}

Fig.~\ref{fig:initial-guess-error} shows the accuracy as a sweep over $r$ for both initial-guess approximations of $\hat b$. As an initial guess, the reused upper bound is exact in the limit $r \to 1$ and its error grows with $r$, while the Lambert $W$ form tightens as $r$ grows. We use the intersection at $r^\ast = \num{3.77}$ as the regime boundary, which minimizes the worst-case error to \qty{3.4}{\percent}.  Furthermore, Table~\ref{tab:initial-guess-iterations} shows that using the analytical guess reduces the average \emph{rtsafe} iterations by \qty{69.2}{\percent} compared to the textbook implementation that uses the bracket center.

\begin{table}[tb]
	\centering
	\caption{Number of iterations and ratio of \emph{rtsafe} comparing the textbook initialization at the bracket midpoint against our analytic guess, evaluated on Flight~1's \num{266} solves. The analytic guess reduces the mean iteration count by \qty{69.2}{\percent} and is never worse on any individual solve.}
	\label{tab:initial-guess-iterations}
	\begin{tabular}{lrrr}
		\toprule
		Initialization & Mean & Median & Max \\
		\midrule
		Midpoint & \num{7.27} & \num{7} & \num{10} \\
		Analytic & \num{2.24} & \num{2} & \num{3} \\
		\midrule
		Ratio & \num{3.25}$\times$ & \num{3.50}$\times$ & \num{3.33}$\times$ \\
		\bottomrule
	\end{tabular}
\end{table}

\subsection{Tolerance and Floating-Point Precision}
\label{sec:tolerance}

Our implementation terminates once the step length falls below a relative tolerance, $\Delta_n < \varepsilon_b\,\hat b_0$ (Sec.~\ref{sec:solver}). Since $\hat b_0 \leq 1.034\,\hat b^\ast$ (Fig.~\ref{fig:initial-guess-error}), this is a relative step of at most $1.034\,\varepsilon_b$. In the quadratic-convergence regime the remaining error is then $\mathcal{O}(\Delta_n^2)$, far below $\varepsilon_b$, and one extra iteration buys about four more decimal digits. At a mean of only \num{2.24} iterations (Table~\ref{tab:initial-guess-iterations}) this extra iteration is expensive.

A threshold below the resolution of the floating-point format cannot be met reliably: near the root, $g$ is dominated by rounding noise and $\hat b$ is only resolved to one unit in the last place. ArduPilot's Lua interpreter uses float32, where the 23-bit mantissa gives a relative resolution of $2^{-23} \approx \num{e-7}$, and the float64 used in Python has a 52-bit mantissa and thus gives a resolution of \num{e-16}. If the threshold is chosen below the resolution, \emph{rtsafe} still terminates, but wastes iterations that do not improve the accuracy.

Table~\ref{tab:tolerance} shows a sweep over different tolerance values. We use $\varepsilon_b = \num{e-4}$, which has sufficient accuracy while ensuring lower iteration counts and avoiding issues with floating-point precision. With this choice, $N_{\mathrm{max}} = \num{28}$ for $r \leq \num{100}$ and never exceeds \num{32} on $(1, r_{\mathrm{max}}]$. In practice the bound is never approached: over a logarithmic sweep of \num{20000} values of $r \in (1, r_{\mathrm{max}}]$ the solver required at most three iterations and never fell back to a bisection step, so the initial guess lands in the quadratic-convergence regime.

\begin{table}[tb]
	\centering
	\caption{Sweep of the relative \emph{rtsafe} step tolerance $\varepsilon_b$ on Flight~1's \num{266} solves, using float64 precision. Errors are the deviation of the endpoint force from the reference setting $\varepsilon_b = \num{e-14}$; the residual is the arc-length error of the recovered catenary. No case failed to converge at any tolerance.}
	\label{tab:tolerance}
	\footnotesize
	\setlength{\tabcolsep}{4pt}
	\sisetup{
		table-format = 1.1e-2,
		table-number-alignment = center
	}
	\begin{tabular}{lSSSS}
		\toprule
		& \multicolumn{2}{c}{Force error [\unit{\newton}]} & \multicolumn{2}{c}{Residual [\unit{\meter}]} \\
		\cmidrule(lr){2-3}\cmidrule(lr){4-5}
		{$\varepsilon_b$} & {Mean} & {Max} & {Mean} & {Max} \\
		\midrule
		\num{e-14} & {---}    & {---}    & 2.5e-14 & 1.1e-13 \\
		\num{e-10} & 1.9e-16  & 7.1e-15  & 2.5e-14 & 1.1e-13 \\
		\num{e-8}  & 1.2e-15  & 1.1e-14  & 2.5e-14 & 1.3e-13 \\
		\num{e-6}  & 1.9e-12  & 1.3e-10  & 1.4e-11 & 9.5e-10 \\
		\num{e-5}  & 2.2e-11  & 1.4e-9   & 1.6e-10 & 1.0e-8  \\
		\num{e-4}  & 2.2e-8   & 7.8e-7   & 1.5e-7  & 5.8e-6  \\
		\num{e-3}  & 2.5e-6   & 7.1e-5   & 1.8e-5  & 5.2e-4  \\
		\bottomrule
	\end{tabular}
\end{table}

\subsection{Runtime Cost of One Catenary Solve}
\label{sec:solve-cost}

\begin{table}[tb]
	\centering
	\caption{Comparison of total solve time in \unit{\nano\second} and ratio between textbook and specialized \emph{rtsafe} implementations, compiled (C, Numba) and interpreted (Python, Lua 5.3). Minimum over \num{25} interleaved batches of \num{20000} solves (C, Numba) or \num{4000} solves (Lua, Python) for an example matching the median of \num{2} iterations, evaluated on a desktop CPU. We observe higher speedup in interpreted languages due to per-dispatch costs and fewer opportunities for compiler optimization.}
	\label{tab:solve-totals}
	\begin{tabular}{lrrrr}
		\toprule
		& C (gcc)  & Numba & Lua & Python \\
		\midrule
		Textbook    & \num{51.3} & \num{51.2} & \num{948} & \num{1431} \\
		Specialized & \num{45.3} & \num{45.9} & \num{725} & \num{1048} \\
		\midrule
		Ratio & \num{1.132}$\times$ & \num{1.115}$\times$ & \num{1.308}$\times$ & \num{1.365}$\times$ \\
		\bottomrule
	\end{tabular}
\end{table}

As shown in Table~\ref{tab:solve-totals}, the specialization gives a speedup of \num{1.132} in C and \num{1.115} with the Numba JIT. On the given example with \num{2} iterations it halves the number of function evaluations from \num{4} to \num{2}, so a larger difference might be expected. In compiled code, however, the compiler and the out-of-order processor reorder and overlap independent operations, so removed work does not translate one-to-one into saved runtime. In interpreted languages every operation incurs a dispatch overhead that dominates its cost and leaves less opportunity for such optimization, so the number of operations translates more directly into runtime, which is consistent with the larger speedups for Python and Lua.

All the optimizations performed by the compiler and the processor pipeline make it difficult to break down the runtime of different components of the algorithm. Therefore, to measure the individual parts, Table~\ref{tab:solve-breakdown} uses Python and Lua, which are interpreted languages, so the total solve time can be assumed to be the sum of the solve times of the parts. This measurement was done by timing each part on its own, with its inputs supplied by variable lookups. Additionally, the variable lookups were timed by themselves to remove their overhead. As expected we see a bigger speedup where the optimized version differs from textbook \emph{rtsafe}. Another interesting observation is that due to the interpreter overhead the function evaluation no longer dominates the runtime.

\begin{table}[tb]
	\centering
	\caption{Breakdown of one catenary solve, textbook vs.\ specialized \emph{rtsafe}, each as a percentage of that language's textbook total (Python: \qty{1431}{\nano\second}, Lua: \qty{948}{\nano\second}). Evaluated for an example matching the median of \num{2} iterations on a desktop CPU. We observe that due to per-dispatch cost the function evaluation does not dominate the runtime.}
	\label{tab:solve-breakdown}
	\begin{tabular}{lrrrr}
		\toprule
		& \multicolumn{2}{c}{Python} & \multicolumn{2}{c}{Lua} \\
		\cmidrule(lr){2-3}\cmidrule(lr){4-5}
		Component & Textbook & Spec. & Textbook & Spec. \\
		\midrule
		Bracket validation    & \qty{22.4}{\percent} & --- & \qty{20.1}{\percent} & --- \\
		Function evaluation   & \qty{22.1}{\percent} & \qty{22.1}{\percent} & \qty{17.6}{\percent} & \qty{17.6}{\percent} \\
		Iteration step         & \qty{19.5}{\percent} & \qty{17.6}{\percent} & \qty{28.3}{\percent} & \qty{26.0}{\percent} \\
		Remainder               & \qty{36.1}{\percent} & \qty{33.5}{\percent} & \qty{34.1}{\percent} & \qty{32.9}{\percent} \\
		\midrule
		Total                   & \qty{100.0}{\percent} & \qty{73.2}{\percent} & \qty{100.0}{\percent} & \qty{76.4}{\percent} \\
		\bottomrule
	\end{tabular}
\end{table}

\subsection{Runtime of the Full Tether Simulation}
\label{sec:runtime}
\begin{figure}[tb]
	\centering
	\includegraphics[width=\columnwidth]{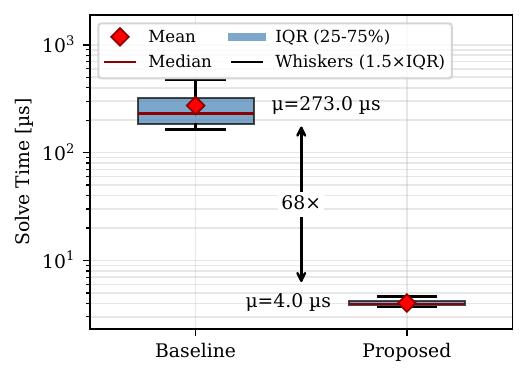}
	\caption{Comparison between the solve time of the previous and proposed method on Flight~1 (\num{170} configurations), evaluated on a desktop CPU running in Python with Numba. Both converge on the same result in all cases and the proposed method shows mean speedup of \num{68} times with a nearly constant runtime. Note the log scale.}
	\label{fig:runtime}
\end{figure}

Figure~\ref{fig:runtime} shows a comparison between the previous and proposed method on Flight~1. Both converge on all \num{170} configurations, and their tension profiles agree to a relative deviation of \num{8.7e-9}. Note that we are comparing against an optimized version of the previous method that is already \num{1.9} times faster than the runtime stated in \cite{beffert_low-latency_2026}.
The proposed formulation solves in \qty{4.0}{\micro\second} on average, against \qty{273.0}{\micro\second} for the baseline: a speedup of \num{68} times, or \num{126} times over the published \qty{0.51}{\milli\second}. The standard deviation drops from \qty{130.5}{\micro\second}, where it depends on how many initial guesses were tried before convergence, to \qty{0.2}{\micro\second}, and the worst case of \qty{4.6}{\micro\second} still beats the fastest baseline solve. For a real-time solver, such a bounded worst case matters as much as the average cost.

\begin{figure}[tb]
	\centering
	\includegraphics[width=\columnwidth]{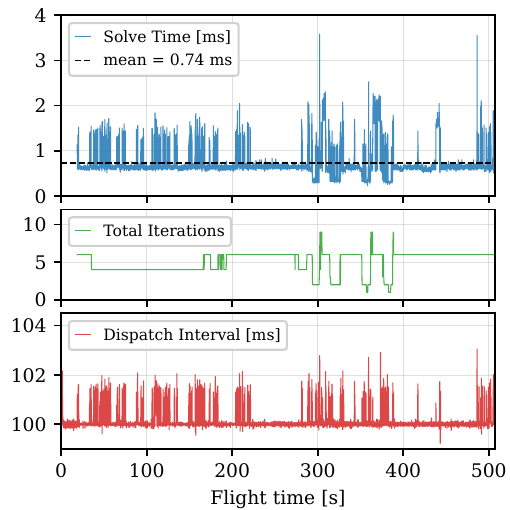}
	\caption{Telemetry from the Lua script running in ArduPilot on the drone flight controller during Flight~2. All \num{4875} configurations converged. Elevated jitter in the dispatch interval (target \qty{10}{\hertz}) coincides with solve time jitter at constant iteration count, indicating interruptions by the real-time operating system.}
	\label{fig:drone-runtime}
\end{figure}

Figure~\ref{fig:drone-runtime} shows the Lua implementation running directly in ArduPilot on the drone flight controller during Flight~2. Every one of its \num{4875} configurations converged, with a mean solve time of \qty{0.74}{\milli\second} and a worst case of \qty{3.58}{\milli\second}, which is inside the \qty{100}{\milli\second} budget of the \qty{10}{\hertz} scheduling loop. This is also reflected by the dispatch interval, which stays within \qty{3.06}{\milli\second} of the \qty{100}{\milli\second} nominal period. This indicates that the script is never aborted or skipped by the scheduler. The elevated jitter in the dispatch interval lines up with the solve time jitter without a corresponding change in iteration count, so it stems from the way Lua is run on ArduPilot rather than from the algorithm: the real-time operating system schedules the Lua interpreter as a low-priority task that can be interrupted by more important processes and resumed afterward.

\section{Conclusion}

In this work, the catenary solved at the core of the quasi-analytical tether model of \cite{beffert_low-latency_2026} is reformulated to a single transcendental equation in the unsigned quantity $\hat b$, with the remaining catenary parameters recovered in closed form. Unlike the shape parameter of the catenary, $\hat b$ is confined to a finite interval, free of singularities, and has a well-behaved derivative throughout its domain. Building on this formulation, a closed-form upper bracket is derived and its tightness proven. Furthermore, we prove that the reduced formulation is both monotonically increasing and convex, thus guaranteeing convergence. 

The problem is solved via a simplified implementation of \emph{rtsafe}, a hybrid root-finding method comprised of Newton-Raphson steps with a fallback to bisection, which gives a guaranteed upper bound to the iteration count. Exploiting the proven monotonicity and convexity, the specialized version omits unnecessary checks while retaining correctness and achieving up to \num{1.37} times speedup over textbook \emph{rtsafe}. Furthermore, a two-regime initial guess was proposed, consisting of the reused upper bound and a Lambert $W$ approximation. It was shown to reduce the mean iteration count by \qty{69.2}{\percent} to \num{2.24} by achieving a worst-case approximation error of \qty{3.4}{\percent}, ensuring fast convergence.

Effectively these improvements result in a solve time of \qty{4.0}{\micro\second} on average and \qty{4.6}{\micro\second} at worst for the full tether model. Compared to an average of \qty{273.0}{\micro\second} with the previous method, this is a \num{68} times speedup. The standard deviation drops from \qty{130.5}{\micro\second} to \qty{0.2}{\micro\second}, making the runtime essentially constant, which is an important factor for real-time use. The two formulations agree to a relative deviation of \num{8.7e-9}; therefore, the correctness claims and experimental validation from \cite{beffert_low-latency_2026} are unaffected. The guarantees cover the inner solve; the unchanged outer drag loop is capped at \num{10} iterations and never exceeded \num{3} in our data, but is not formally guaranteed to converge.

Because the relative tolerance is chosen against the platform's floating-point resolution, all quantities are evaluated free of catastrophic cancellation, and the custom \emph{rtsafe} needs only basic math operations, the solver ports easily to different platforms. This was demonstrated by writing a Lua implementation that runs directly in ArduPilot on the drone's flight controller. It achieved a mean runtime of \qty{0.74}{\milli\second} during flight, successfully solving all cases and staying within the budget of the scheduling loop, demonstrating its real-time capability.

The assumptions of the underlying model are unchanged: the tether is inextensible with uniform mass and drag distribution along its length, and the configuration is quasi-static and planar. While we implement the 2D case, under the given assumptions the solution stays planar in 3D \cite{chakrabarti_catenaries_2016}, which means the proposed improvements extend to that as well. Cases where higher physical fidelity is required remain the domain of the numerical model of \cite{beffert_low-latency_2026}, so the two methods continue to complement each other. Future work includes applying the model within online wind estimation, model predictive control, and software-in-the-loop simulation, where its strength of bounded worst-case runtime provides the most benefit. 

The implementation will be made publicly available upon publication of this work.


%%%%%%%%%%%%%%%%%%%%%%%%%%%%%%%%%%%%%%%%%%%%%%%%%%%%%%%%%%%%%%%%%%%%%%%%%%%%%%%%

%%%%%%%%%%%%%%%%%%%%%%%%%%%%%%%%%%%%%%%%%%%%%%%%%%%%%%%%%%%%%%%%%%%%%%%%%%%%%%%%

%%%%%%%%%%%%%%%%%%%%%%%%%%%%%%%%%%%%%%%%%%%%%%%%%%%%%%%%%%%%%%%%%%%%%%%%%%%%%%%%

%%%%%%%%%%%%%%%%%%%%%%%%%%%%%%%%%%%%%%%%%%%%%%%%%%%%%%%%%%%%%%%%%%%%%%%%%%%%%%%%

\bibliographystyle{IEEEtranDOI}
\bibliography{references}

\section*{APPENDIX}

\makeatletter

\def\thesubsection{\Alph{subsection}}
\def\thesubsubsection{\thesubsection.\arabic{subsubsection}}

\def\thesubsectiondis{\Alph{subsection}}
\def\thesubsubsectiondis{\Alph{subsection}.\arabic{subsubsection}}

\def\subsection{\@startsection{subsection}{2}{\z@}%
	{1.5ex plus 1.5ex minus 0.5ex}%
	{0.7ex plus .5ex minus 0ex}%
	{\normalfont\normalsize}}

\def\subsubsection{\@startsection{subsubsection}{3}{\z@}%
	{0ex plus 0.1ex minus 0.1ex}%
	{0.7ex plus 0.2ex}%
	{\normalfont\normalsize}}

\makeatother

\subsection{Catenary Reduction to One Unknown}\label{analytical-appendix}

The curve is a catenary of the form 
\begin{equation*}
y(x) = a \cosh\big((x - x_0)/a\big) + y_0.
\end{equation*}
Given two endpoints $p_1=(x_1, y_1)$, $p_2=(x_2, y_2)$ on the curve and
a prescribed arc length $L$ between them, the goal is to determine the three unknowns: the scale parameter $a > 0$ and the vertex coordinates $(x_0, y_0)$.

\subsubsection{Substitution of Normalized Coordinates}
The arc length $L$ is the sum of the signed arc lengths from the vertex to $p_1$ and $p_2$,
\begin{equation*}
	L = a \sinh\left(\frac{x_2-x_0}{a}\right)-a\sinh\left(\frac{x_1-x_0}{a}\right).
\end{equation*}
Introducing the normalized coordinates $u_i := (x_i - x_0)/a$, this becomes $L = a\,\big(\sinh u_2 - \sinh u_1\big)$. 
Similarly, the vertical separation is
\begin{equation*}
	\Delta y = y_2 - y_1 = a\,\big(\cosh u_2 - \cosh u_1\big).
\end{equation*}

\subsubsection{Elimination of the Midpoint Parameter}
Applying the addition identities gives
\begin{align*}
	L &=  2a \, \cosh \left(\frac{u_1 + u_2}{2}\right)  \sinh \left(\frac{u_2 - u_1}{2}\right), \\
	\Delta y &= 2a \, \sinh \left(\frac{u_1 + u_2}{2}\right)  \sinh \left(\frac{u_2 - u_1}{2}\right).
\end{align*}
With the substitutions
\begin{align*}
	m &:= \frac{u_1 + u_2}{2},\\
	b &:= \frac{u_2 - u_1}{2} = \frac{x_2 - x_1}{2a} = \frac{\Delta x}{2a},
\end{align*}
these read
\begin{align}
	L      &= 2a \, \cosh m \: \sinh b, \label{eq:L2}\\
	\Delta y &= 2a \, \sinh m \: \sinh b. \label{eq:dy2}
\end{align}
Squaring and subtracting eliminates $m$ via the identity $\cosh^2 m - \sinh^2 m = 1$, leaving
\begin{equation}
	L^2 - \Delta y^2 = 4a^2 \sinh^2 b.
	\label{eq:intermediate_appendix}
\end{equation}

\subsubsection{Reduction to a Single Unknown}\label{reduction-analytical-appendix}
Substituting $a := \frac{\Delta x}{2b}$
turns the previous relation into
\begin{equation*}
	L^2 - \Delta y^2 = 4\left(\frac{\Delta x}{2b}\right)^2 \sinh^2 b
	= \left(\frac{\Delta x}{b}\right)^2 \sinh^2 b.
\end{equation*}
Taking the positive square root ($\hat b$ denotes the unsigned $b$) yields a single transcendental equation in one unknown,
\begin{equation}
	\boxed{\;g(\hat b) = |\Delta x| \: \frac{\sinh \hat b}{\hat b \rule{0pt}{2.4ex}} - \sqrt{L^2 - \Delta y^2} = 0.}
	\label{eq:g_of_b_appendix}
\end{equation}
For nearly taut tethers, $\sqrt{L^2 - \Delta y^2}$ and $|\Delta x| \sinh(\hat b)/\hat b$ both approach $|\Delta x|$, so their difference in \eqref{eq:g_of_b_appendix} suffers from catastrophic cancellation. We instead solve the equivalent, cancellation-free form
\begin{equation}
	\frac{\sinh \hat b}{\hat b} - 1 - \frac{(L-D)(L+D)}{|\Delta x|\,(C + |\Delta x|)} = 0,
	\label{eq:g_cancel_free}
\end{equation}
obtained from dividing $g(\hat b)$ by $|\Delta x|$ and using $C^2 - \Delta x^2 = (L-D)(L+D)$ with $C = \sqrt{(L-|\Delta y|)(L+|\Delta y|)}$ and $D = \sqrt{\Delta x^2 + \Delta y^2}$.

\vspace{1ex}
\subsubsection{Recovery of the Remaining Parameters}\label{recovery-analytical-appendix}
Once \eqref{eq:g_of_b_appendix} is solved for $\hat b$, the original three unknowns follow in closed form. Because of the positive square root, \eqref{eq:g_of_b_appendix} only gives $\hat b$ as the magnitude of $b$, so the sign must be recovered from $\Delta x$:
\begin{equation}
	a = \frac{\Delta x}{\operatorname{sgn}(\Delta x)\:2\hat b \rule{0pt}{2.4ex}}=\frac{|\Delta x|}{2\hat b \rule{0pt}{2.4ex}}.
\end{equation}
Dividing \eqref{eq:dy2} by \eqref{eq:L2} gives $\tanh (m) = \Delta y/L$. Since the algebraic $L$ in this ratio can be negative, using the prescribed positive $L$ instead yields the sign-normalized $\hat m$. Evaluated directly, $\operatorname{atanh}(\Delta y/L)$ is ill-conditioned as $|\Delta y| \to L$. Instead we use the identical but well-conditioned form:
\begin{equation*}
	\hat m = \operatorname{asinh}\!\left(\frac{\Delta y}{C}\right).
\end{equation*}
With $u_1 = m - b$ and the recovered sign,
\begin{align}
	u_1 &= \operatorname{sgn}(\Delta x)\,(\hat m - \hat b),\\
	x_0 &= x_1 - a\,u_1, \\
	y_0 &= y_1 - a\,\cosh u_1.
\end{align}
This completes the recovery of the unknowns $(a, x_0, y_0)$.

\vspace{1ex}
\subsection{Monotonicity and Convexity of $g(\hat b)$}\label{monotonicity-appendix}

We can write $g(\hat b) = |\Delta x|\, f(\hat b) - C$ with constants $C$ and $|\Delta x|$, where $f(\hat b) := \sinh(\hat b)/\hat b$. Then the monotonicity and convexity of $g$ on $(0,\infty)$ follow directly from those of $f$.

Dividing the Taylor series of $\sinh \hat b$ term-by-term by $\hat b$ gives
\begin{equation*}
	\begin{aligned}
	f(\hat b) &= \frac{\sinh \hat b}{\hat b} = 1 + \frac{\hat b^2}{3!} + \frac{\hat b^4}{5!} + \cdots =\sum_{n=0}^{\infty} \frac{\hat b^{\,2n}}{(2n+1)!}\\
	&=\sum_{n=0}^{\infty} c_n \hat b^{2n}, \qquad c_n := \frac{1}{(2n+1)!} > 0
	\end{aligned}
\end{equation*}
Differentiating termwise,
\begin{align*}
	f'(\hat b) &= \sum_{n=1}^{\infty} 2n\,c_n\, \hat b^{2n-1}, \\
	f''(\hat b) &= \sum_{n=1}^{\infty} 2n(2n-1)\,c_n\, \hat b^{2n-2}.
\end{align*}
Every coefficient $c_n$ is positive, and for $\hat b > 0$ every power $\hat b^{2n-1}$ and $\hat b^{2n-2}$ is nonnegative, so both series consist entirely of nonnegative terms:
\begin{equation}
	f'(\hat b) > 0 \ \text{ and } \ f''(\hat b) > 0 \qquad \text{for all } \hat b > 0.
\end{equation}
Hence $f$, and therefore $g$, is strictly increasing and strictly convex on $(0,\infty)$.

\subsection{Correctness of the Simplified Bracket Check}\label{low-bracket-appendix}
Our implementation omits the check from standard \emph{rtsafe} that ensures a Newton step never goes below the lower bracket bound. We show that the lower bracket is never violated, given the convexity and monotonicity established in \refappendix{monotonicity-appendix}.
\paragraph{Lemma} Let $g$ be convex and strictly increasing on $(0,\infty)$, and let $\hat b^\ast \in (0,\infty)$ be its root. For any $\hat b_n > 0$, the Newton step $\hat b_{n+1} = \hat b_n - g(\hat b_n)/g'(\hat b_n)$ satisfies $\hat b_{n+1} \geq \hat b^\ast$.
\paragraph{Proof} Convexity means the tangent line at $\hat b_n$ lies below $g$ everywhere:
\begin{equation*}
	g(x) \geq g(\hat b_n) + g'(\hat b_n)(x - \hat b_n) \qquad \text{for all } x > 0.
\end{equation*}
Evaluating at $x = \hat b^\ast$ and using $g(\hat b^\ast) = 0$,
\begin{equation*}
	0 \geq g(\hat b_n) + g'(\hat b_n)(\hat b^\ast - \hat b_n).
\end{equation*}
Since $g'(\hat b_n) > 0$ by monotonicity, dividing by $g'(\hat b_n)$ preserves the inequality and we can rearrange to:
\begin{equation}
	\hat b^\ast \leq \hat b_n - \frac{g(\hat b_n)}{g'(\hat b_n)} = \hat b_{n+1}.
\end{equation}
Since the bracket invariant guarantees $\hat b^\ast \geq \hat b_{\mathrm{lo}}$ at every step, the lemma gives $\hat b_{n+1} \geq \hat b^\ast \geq \hat b_{\mathrm{lo}}$ for any positive iterate. The Newton step can therefore never undershoot past $\hat b_{\mathrm{lo}}$.

\vspace{1ex}
\subsection{Existence of a Unique Root}\label{existence-appendix}
Since $g$ is strictly increasing on $(0,\infty)$ (\refappendix{monotonicity-appendix}), it has at most one root in the interval. It remains to show that $g$ changes sign across the domain, so a root exists.

As $\hat b \to 0^+$, $f(\hat b) \to 1$, so $g(0^+) = |\Delta x| - C$. By the first well-posedness guard of Sec.~\ref{sec:guards}, $L^2 > \Delta x^2 + \Delta y^2$, which gives $C^2 = L^2 - \Delta y^2 > \Delta x^2$ and thus $C > |\Delta x|$. Consequently $g(0^+) < 0$. Since $f(\hat b) \to \infty$ as $\hat b \to \infty$, so does $g(\hat b)$. There must therefore be a zero crossing in the interval $(0,\infty)$. Combined with the injectivity established above, this shows that the root exists and is unique.

\vspace{1ex}
\subsection{Upper Bound for $\hat b$}\label{bounds-appendix}
We seek a closed-form equation for $\hat b_{\mathrm{hi}}$ with $g(\hat b_{\mathrm{hi}}) \geq 0$, i.e., $f(\hat b_{\mathrm{hi}}) \geq r$, where $r := \sqrt{L^2-\Delta y^2}/|\Delta x|$. Recall from \refappendix{existence-appendix} that well-posedness requires $C > |\Delta x|$, thus $r>1$, which is used throughout this derivation.

Since every coefficient $c_n$ in the series $f(\hat b)~=~\sum_{n=0}^{\infty} c_n \hat b^{2n}$ is positive (\refappendix{monotonicity-appendix}), truncating the series only discards positive values. Therefore any partial sum is a lower bound on $f$ for $\hat b \geq 0$. Truncating after the quartic term gives
\begin{equation}\label{eq:taylor-approximation}
	f(\hat b) \; \geq \; h(\hat b) := 1 + \frac{\hat b^2}{6} + \frac{\hat b^4}{120} \qquad \text{for all } \hat b \geq 0.
\end{equation}
We use this approximation to find $\hat b_{\mathrm{hi}}$, so we solve $h(\hat b) = r$. Substituting $w := \hat b^2$ turns this into a quadratic,
\begin{equation*}
	\frac{w^2}{120} + \frac{w}{6} + (1-r) = 0 \quad\Longleftrightarrow\quad w^2 + 20w + 120(1-r) = 0,
\end{equation*}
with solution
\begin{equation*}
	w = -10 + \sqrt{100 + 120(r-1)}.
\end{equation*}
The other root is negative for $r>1$ and hence discarded, since $w=\hat b^2 \geq 0$ is required. As $r \to 1$ this form subtracts two nearly equal numbers, so we multiply by the conjugate to obtain an algebraically identical form free of cancellation. Taking the positive square root gives the closed-form bound
\begin{equation}\label{eq:upper-bound-appendix}
	\boxed{\;\hat b_{\mathrm{hi}} = \sqrt{w}=\sqrt{\frac{120(r-1)}{10 + \sqrt{100 + 120(r-1)}}}.}
\end{equation}
In practice, we scale this by a factor of $(1+\num{e-6})$ to guard against floating-point round-off, ensuring $g(\hat b_{\mathrm{hi}}) > 0$ strictly rather than borderline.

\vspace{1ex} 
\subsubsection{Tightness of the Bound}\label{bounds-tightness-appendix} 
As $r \to 1^+$ the root $\hat b^\ast \to 0$, so the discarded higher-order terms vanish quickly and the bracket becomes exact. The relevant regime is therefore large $r$, where the quartic term of \eqref{eq:taylor-approximation} dominates and gives $\hat b_{\mathrm{hi}} \sim (120\,r)^{1/4}$, while the root grows only logarithmically, $\hat b^\ast \sim \ln(2r)$ (\refappendix{initial-guess-large-appendix}). The bracket thus loosens as $r^{1/4}/\ln r$. Because bisection to a relative tolerance of $\varepsilon_b$ needs $\log_2\!\big(\hat b_{\mathrm{hi}}/(\varepsilon_b \hat b_0)\big)$ steps, the penalty compared to a perfect bracket is $\log_2(\hat b_{\mathrm{hi}}/\hat b^\ast)$, which is below half a step for $r \leq \num{100}$ and below three steps even at $r_{\mathrm{max}}$.

\vspace{1ex}
\subsection{Initial Guess for the Root of $g(\hat b)$}\label{initial-guess-appendix}
As in \refappendix{bounds-appendix}, write
$r := \sqrt{L^2 - \Delta y^2}/|\Delta x| > 1$, so that the root satisfies $f(\hat b) = r$ with $f(\hat b) = \sinh(\hat b)/\hat b$. For small $\hat b$, the upper bound $\hat b_{\mathrm{hi}}$ \eqref{eq:upper-bound-appendix} is itself a tight estimate of the root and is reused directly as the initial guess. For large $\hat b$ we derive a closed-form approximation below.

\vspace{1ex}
\subsubsection{Large-$\hat b$ Regime}\label{initial-guess-large-appendix}
For $\hat b \gg 1$, the $e^{-\hat b}$ term in $\sinh \hat b/\hat b = (e^{\hat b}-e^{-\hat b})/2\hat b$ is exponentially negligible, giving the approximation
\begin{equation*}
	f(\hat b) \approx \frac{e^{\hat b}}{2\hat b}.
\end{equation*}
Solving $f(\hat b) = r$ for $\hat b$,
\begin{equation*}
	\frac{e^{\hat b}}{2\hat b} = r
	\quad\Longleftrightarrow\quad
	(-\hat b)\, e^{-\hat b} = -\frac{1}{2r},
\end{equation*}
matches the defining relation $z = W(z)\,e^{W(z)}$ of the Lambert $W$ function, with $z = -1/(2r)$ and $W(z) = -\hat b$. 
Recall from \refappendix{existence-appendix} that well-posedness gives $r>1$ and therefore $z \in \left(-\tfrac12, 0\right) \subset \left(-\tfrac1e, 0\right)$, which is the interval on which $W$ has two real branches. Since $W_0(0)=0$, the principal branch yields $\hat b\to0$ as $z\to0^-$ and is therefore inconsistent with the large-$\hat b$ regime. We instead take the $W_{-1}$ branch, which gives $\hat b = -W_{-1}(-1/(2r))$. Its standard asymptotic expansion as $z\to0^-$, with $\Lambda := \ln(2r)$ \cite{corless_lambertw_1996}, gives
\begin{equation}
	\boxed{\;\hat b \approx \Lambda + \ln \Lambda + \ln \Lambda/\Lambda\;} \qquad (r \text{ large}).
	\label{eq:guess_large_appendix}
\end{equation}

\end{document}